\documentclass[journal]{IEEEtran}
\usepackage{amsmath,amsfonts}
\usepackage{algorithmic}
\usepackage{algorithm}
\usepackage{array}
\usepackage[caption=false,font=normalsize,labelfont=sf,textfont=sf]{subfig}
\usepackage{textcomp}
\usepackage{stfloats}
\usepackage{url}
\usepackage{booktabs}
\usepackage{verbatim}
\usepackage{graphicx}
\usepackage{balance}
\usepackage{cite}
\usepackage{color}
\usepackage{mathrsfs}
\begin{document}

\title{AIA-UltraNeRF: Acoustic-Impedance-Aware Neural Radiance Field with Hash Encodings for Robotic Ultrasound Reconstruction and Localization}

\author{Shuai Zhang$^\dagger$, Jingsong Mu$^\dagger$, Cancan Zhao, Leiqi Tian, Zhijun Xing, Bo Ouyang, Xiang Li
\thanks{This work was supported in part by the Young Scientists Fund of the National Natural Science Foundation of China (Grant No. 52305018), in part by the National Key Research and Development Program of China (Grant No. 2023YFB4706000), in part by the Joint Funds Program of the National Natural Science Foundation of China (Grant No. U21A20517), in part by the Basic Science Centre Program of the National Natural Science Foundation of China (Grant No. 72188101). (Corresponding author: Bo Ouyang)}
\thanks{Shuai Zhang, Cancan Zhao, and Bo Ouyang are with the School of Management, Hefei University of Technology, Hefei 230009, China (e-mails: zshuai\_9508@163.com; cz962163@gmail.com; boouyang@hfut.edu.cn)}
\thanks{Jingsong Mu is with the First Affiliated Hospital of University of Science and Technology of China (USTC), Hefei 230001, China (email: jsmu@ustc.edu.cn).}
\thanks{Leiqi Tian is with the Department of Ultrasound Diagnosis, the Second Xiangya Hospital of Central South University, Changsha 410011, China (email: tianleiqi@csu.edu.cn).}
\thanks{Zhijun Xing is with the Intelligent Ultrasound Division of VINNO Technology Co., Ltd., Suzhou 215000, China (email: justin.xing@vinno.com).}
\thanks{Xiang Li is with the Department of Automation, Tsinghua University, Beijing 100084, China (email: xiangli@tsinghua.edu.cn).}
\thanks{$\dagger$ These authors contributed equally.}
}

\markboth{Journal of \LaTeX\ Class Files,~Vol.~14, No.~8, August~2021}%
{Shell \MakeLowercase{\textit{et al.}}: A Sample Article Using IEEEtran.cls for IEEE Journals}


\maketitle
\begin{abstract}
Neural radiance field (NeRF) is a promising approach for reconstruction and new view synthesis. However, previous NeRF-based reconstruction methods overlook the critical role of acoustic impedance in ultrasound imaging. Localization methods face challenges related to local minima due to the selection of initial poses. In this study, we design a robotic ultrasound system (RUSS) with an acoustic-impedance-aware ultrasound NeRF (AIA-UltraNeRF) to decouple the scanning and diagnostic processes. Specifically, AIA-UltraNeRF models a continuous function of hash-encoded spatial coordinates for the 3D ultrasound map, allowing for the storage of acoustic impedance without dense sampling. This approach accelerates both reconstruction and inference speeds. We then propose a dual-supervised network that leverages teacher and student models to hash-encode the rendered ultrasound images from the reconstructed map. AIA-UltraNeRF retrieves the most similar hash values without the need to render images again, providing an offline initial image position for localization. Moreover, we develop a RUSS with a spherical remote center of motion mechanism to hold the probe, implementing operator-independent scanning modes that separate image acquisition from diagnostic workflows. Experimental results on a phantom and human subjects demonstrate the effectiveness of acoustic impedance in implicitly characterizing the color of ultrasound images. AIA-UltraNeRF achieves both reconstruction and localization with inference speeds that are 9.9× faster than those of vanilla NeRF. 
\end{abstract}

\begin{IEEEkeywords}
Ultrasound reconstruction, ultrasound localization, neural radiance fields, acoustic impedance, hash encoding.
\end{IEEEkeywords}

\section{Introduction}
\IEEEPARstart{U}{ltrasound} is a mainstay of contemporary medical imaging for various diagnostic applications due to its non-invasive nature, high safety, and cost-effectiveness. To obtain the standard plane, sonographers mentally construct a 3D map of the patient's organs and gradually adjust the position of the probe based on the anatomical features observed in the ultrasound images \cite{enriquez2014introduction}. However, sonographers require extensive and deliberate practice to develop this skill, in contrast to computed tomography (CT) scanning. Integrating robotic and deep learning technologies into ultrasound systems enhances the precision and repeatability of diagnostic procedures while simultaneously alleviating musculoskeletal pain and reducing the workload of sonographers \cite{si2024design}.
Despite this advancement, previous robotic ultrasound (RUSS) studies primarily focus on normal ultrasound probe positioning and autonomous scanning with manual features \cite{10399878, 9444419}. Few studies present robotic ultrasound systems that decouple the scanning and diagnostic processes, similar to how CT localizes the standard plane based on reconstructed ultrasound maps.
\begin{figure}[!t]
\centering
\includegraphics[width=3.5in]{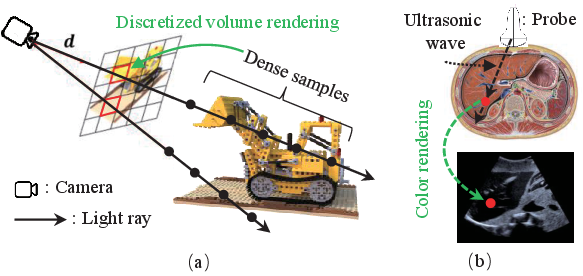}
\caption{The difference between camera and ultrasound imaging. (a) Camera imaging relies on dense samples of light rays to render color. (b) In ultrasound imaging, color is mapped by acoustic impedance at the corresponding position.}
\label{fig1}
\end{figure}

Previous approaches to ultrasound reconstruction, including voxel/pixel-based and learning-based methods, while effective in 3D reconstruction, struggle to accurately render images from new views \cite{guo2022ultrasound, li2023long}. The implicit neural representation (INR) technique, such as NeRF, maps spatial coordinates to target attributes (e.g., color and density) instead of explicitly storing discrete data points, thereby offering an innovative method for reconstructing 3D ultrasound from 2D scans \cite{wysocki2024ultra, mildenhall2021nerf}. NeRF employs a multilayer perception (MLP) to continuously reconstruct scenes, synthesizing realistic images from new views compared to previous methods that utilize convolutional neural networks. This breakthrough is particularly significant when combined with hash grid-based reconstruction techniques, which leverage hash encoding to map inputs into compact representations for rapid feature retrieval, seamlessly aligning with the requirements of medical imaging \cite{mueller2022instant, xu2023nesvor, 11128742}. However, NeRF-based reconstruction in natural scenes relies on dense sampling to model the intersections between rays and objects. Previous approaches have directly applied the dense sampling strategy, overlooking fundamental principles of ultrasound imaging, which results in slower reconstruction and inference speeds \cite{zhang2025ultra}.

The color variations in ultrasound images arise from differences in acoustic impedance, which measures a medium's resistance to sound wave propagation \cite{mahesh2013essential, shung2012principles}. These differences influence how ultrasonic waves propagate and reflect between tissues, resulting in the brightness variations observed in B-mode images \cite{patey2021physics}. Consequently, the contrast in ultrasound imaging is not merely a function of intensity; rather, it is a direct consequence of variations in acoustic impedance that delineate tissue boundaries and highlight abnormalities such as tumors or cysts. Existing NeRF-based methods primarily depend on dense spatial sampling and intensity-based rendering rather than explicitly modeling acoustic impedance, as demonstrated in Fig.~\ref{fig1}. Modeling acoustic impedance at each spatial point without dense sampling can enhance reconstruction efficiency and improve inference speed. 

The iterative localization method, such as iNeRF, relies on the initial pose of the probe and the features of the prominent image. This reliance renders it susceptible to local optima due to significant variations in the ultrasound images \cite{yen2021inerf}. The NeRF-based reconstructed 3D ultrasound map enables the rendering of new-view ultrasound images. This capability presents a novel opportunity for the localization and navigation of ultrasound image planes. By extracting features from these rendered images, encoded through neural networks, it is possible to ascertain the probe's pose from the image code. The localization procedure is transformed into an image retrieval process, driven by scoring the alignment of these features with the target anatomical structures \cite{fang2021deep}. This approach circumvents the need for initial pose selection and eliminates the necessity of image rendering at each step, thereby providing offline localization that determines the pose of the image. This is a critical prerequisite for subsequent online localization refinement in RUSS. 

Based on the analysis presented above, this paper proposes AIA-UltraNeRF and introduces a novel scanning mode for RUSS, which potentially enables the decoupling of scanning and diagnostic workflows. AIA-UltraNeRF first leverages multi-resolution hash grids to encode acoustic impedance, allowing for more efficient and accurate ultrasound reconstruction. Specifically, the features of sampled points are stored as learnable feature vectors based on spatial coordinates via the hash function, enabling the aggregation of each point's acoustic impedance characteristics through trilinear interpolation. Rather than directly rendering colors based on ray accumulation, the final color is rendered by combining acoustic impedance and ultrasonic wave direction through a tiny MLP. Beyond reconstruction, AIA-UltraNeRF also facilitates efficient ultrasound localization. Once the 3D ultrasound volume is reconstructed, rendered images from different views are hash-encoded, establishing a direct mapping between probe poses and hash codes. Localization achieved by matching the shortest distance in the hash space enables offline initial pose selection for subsequent online refinement, thus eliminating the need for iterative rendering at each step. A dual-supervised network is introduced to ensure robust feature extraction for hash encoding. AIA-UltraNeRF employs a teacher-student network architecture where parameters are shared, and both models are trained concurrently, thereby mitigating sensitivity to anatomical variations. To support the scanning mode, a spherical remote center of motion (RCM) mounted on a robotic arm is designed for precise 3D ultrasound reconstruction and image plane localization (URIPL) conducted on a phantom and ten participants. Our approach enables sonographers to navigate the reconstructed ultrasound volume and efficiently retrieve the desired image plane, similar to how CT and MRI scans are analyzed after image acquisition. This scanning mode reduces operator-dependent variability by decoupling ultrasound localization from real-time diagnostic workflows.
The contributions of this paper are summarized as follows:
\begin{itemize}
  \item To the best of our knowledge, this is the first use of acoustic impedance in NeRF-based ultrasound reconstruction. AIA-UltraNeRF encodes the acoustic impedance of relevant tissues using hash grids, which eliminates dense sampling and speeds up the rendering process.
  \item We present a dual-supervised network leveraging the teacher and student model to hash encode the rendered images from the reconstructed map, which offers offline initial pose selection through retrieval for localization and avoids repetitive rendering. 
  \item We propose a novel autonomous ultrasound scanning mode based on the designed AIA-UltraNeRF and image encoding method, which facilitates the separation of ultrasound scanning and diagnosis processes. Experiments on phantom and human subjects demonstrate the feasibility of the proposed mode.  
\end{itemize}

The remainder of this paper is organized as follows: Section \ref{chapter 2} introduces related works. Section \ref{chapter 4} describes AIA-UltraNeRF, detailing both ultrasound reconstruction and ultrasound localization. Section \ref{chapter 8} presents the simulations and experiments, while Section \ref{chapter 9} concludes the paper.

\section{Related Works}\label{chapter 2}
\subsection{Robotic Ultrasound Systems}
RUSS holds the probe on the surface while maintaining image quality to acquire diagnostic images \cite{9866020}. Jiang et al. propose a RUSS that leverages contact force estimation to optimize acoustic propagation, thereby enhancing image quality during ultrasound acquisition \cite{jiang2020automatic}. Another RUSS, equipped with evenly distributed distance sensors, facilitates the automatic alignment of the ultrasound probe during acquisition \cite{ma2022see}. However, these systems provide a scanning mode that aligns the probe perpendicular to the skin surface, which is suitable for imaging structures such as blood vessels or the aorta. To extend beyond perpendicular alignment, a visual servoing controller is utilized to guide the manipulator, improving image quality and accurately localizing lesion areas \cite{chatelain2017confidence}. Nevertheless, this approach has limitations when localizing across varying images. A lightweight and portable RUSS design with a cable-sheath mechanism enables tilt control for various imaging tasks \cite{ning2024cable}. Yet, this method requires patients to wear specific equipment, which can increase their burden. Additionally, parallel mechanism-based RUSS has been developed with an RCM constraint, but the parallel mechanism has a limited workspace and occupies substantial physical space \cite{zhang2024development}. In contrast, we propose a RUSS with a large workspace and an autonomous mode that supports the separation of ultrasound scanning and diagnostic processes. The RUSS is equipped with a spherical RCM mechanism that enables rotation and multi-angle tilt scanning, offering efficient robotic ultrasound image acquisition and subsequent localization tasks.

\subsection{INR for Ultrasound Reconstruction}
Recent advancements in INR, such as NeRF, have enabled the reconstruction of detailed 3D ultrasound maps from 2D scans \cite{zhang2025ultra}. NeRF employs MLPs to create a continuous volumetric representation, and some adaptations have incorporated physically grounded ray-tracing models to account for ultrasound-specific properties \cite{wysocki2024ultra}. Yeung et al. propose a method for conducting 3D ultrasound reconstruction by utilizing a set of 2D scans along with their estimated 3D locations as input \cite{yeung2021implicitvol, yeung2024sensorless}. 
RUSS facilitates access to precise locations through robotic tracking, enhancing volumetric reconstruction by compensating for breathing motion in 3D vascular ultrasound via INR \cite{velikova2023implicit}. However, these methods primarily focus on intensity-based rendering while neglecting the crucial role of acoustic impedance, which is essential for accurate tissue differentiation in ultrasound imaging. This oversight creates challenges in distinguishing between tissues with similar intensities but distinct acoustic impedances, leading to limited tissue contrast.

Current research aims to adapt INR for real-time ultrasound applications by incorporating computational techniques and plane-based representations to address challenges in training efficiency \cite{chen2022tensorf, eid2024rapidvol}. However, only the intensities of the images are rendered, which limits the ability to differentiate between tissues with varying acoustic impedances. In ultrasound imaging, acoustic impedance influences the reflection of an ultrasound wave when it encounters a tissue boundary \cite{mahesh2013essential, shung2012principles}. We model the 3D ultrasound map as an acoustic-impedance-aware field without dense sampling, rendering color by aggregating the acoustic impedance at each point, thereby enhancing tissue differentiation and improving computational efficiency.

\subsection{Ultrasound Localization from Reconstructed Map}
Leveraging a precisely reconstructed ultrasound map opens new avenues for achieving high-accuracy ultrasound localization, overcoming the limitations of sequential probe placement for individual imaging. INeRF and its variants represent pioneering methods for ultrasound localization within such a reconstructed framework, allowing for spatially accurate alignment between the probe and the image \cite{yen2021inerf, bortolon2024iffnerf, lin2023parallel}. Despite these advancements, these methods often require precise initial pose estimations and repetitive rendering steps. Furthermore, the convergence process is hindered by the limited distinction of features in ultrasound images, increasing the risk of settling into local minima during iterative alignment.

An alternative approach that does not rely on prior pose knowledge is image retrieval. This method aligns image features with corresponding pose information to facilitate effective ultrasound localization. Deep hashing methods, which convert high-dimensional image features into compact binary codes, have demonstrated potential for improving retrieval efficiency in medical imaging \cite{fang2021deep, bedi2021mean}. However, the inherently low contrast and high noise levels of ultrasound images make feature extraction challenging. To address this issue, some approaches have integrated distillation learning into deep hashing frameworks, utilizing teacher-student model dynamics to capture more distinctive and robust features \cite{shao2021deep, jang2022deep}. We introduce a dual-supervised network that leverages both teacher and student models for robust feature extraction to achieve ultrasound localization. This network provides an offline initial localization method to establish a foundation for subsequent online probe adjustments. 


\begin{figure*}[!t]
\centering
\includegraphics[width=6.7in]{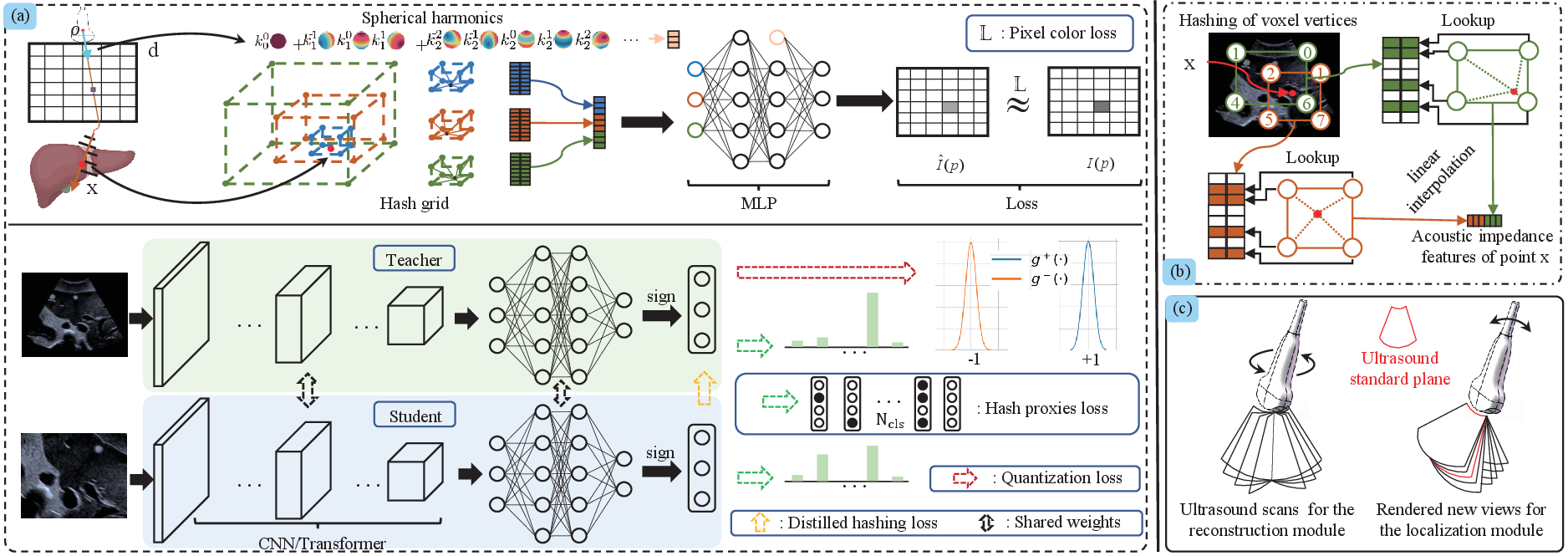} 
\caption{(a) An overview of the AIA-UltraNeRF framework. The top panel shows the reconstruction module, where hash grids and spherical harmonics encode acoustic impedance and wave direction for NeRF-based ultrasound reconstruction, supervised by pixel-level loss. The bottom panel illustrates the localization module, where a teacher-student network learns hash codes of rendered images using multiple loss terms for ultrasound localization. (b) The encoding process of acoustic impedance by hash grids. (c) Ultrasound scans for the reconstruction module and the rendered new views with the standard plane.
}
\label{fig2}
\end{figure*}

\section{AIA-UltraNeRF}\label{chapter 4}
\subsection{Ultrasound Scanning Mode}\label{chapter 5-1}

Ultrasound imaging, in combination with RUSS and NeRF, can establish an innovative ultrasound scanning modality that separates scanning from diagnostics, similar to CT. This mode utilizes RUSS for online image acquisition and NeRF for offline ultrasound reconstruction, enabling the synthesis of new view ultrasound images within the same area. Additionally, it can assist sonographers in localizing the standard plane within the reconstructed map, as illustrated in Fig.~\ref{fig2}. Consequently, ultrasound images can be acquired with remote guidance from experienced sonographers, facilitating diagnosis or reexamination at any time. 

In the online phase, 2D images are captured as \( I_{2D} \in \mathbb{R}^{H \times W} \), where \( H \) and \( W \) represent the height and width of the image. The probe's trajectory during acquisition can be described as:
\begin{align}
\mathbf{T}(t) = \{(x_p, y_p, z_p, \alpha_p, \beta_p, \gamma_p)\},
\end{align}
where $t$ denotes the time along the probe's scanning trajectory, $\mathbf{T}(t)$ represents the 6-DoF pose of the ultrasound probe at that moment, \( (x_p, y_p, z_p) \) is the probe's position, and \( (\alpha_p, \beta_p, \gamma_p) \) is the probe's direction. Following data acquisition, multi-view images \( \{I_{2D}^i\}_{i=1}^N \) (with $N=72$ for the phantom and $N=37$ for each individual) are processed by AIA-UltraNeRF to reconstruct a 3D volume \( V(\mathbf{x})\) described in~\ref{chapter 4-2}. Each point \(\mathbf{x} = [x, y, z]^T \in \mathbb{R}^3 \) in the volume is parameterized by hash grids with acoustic impedance to decode the density $\sigma$ and the color $\textbf{c}$:
\begin{align}
V(\mathbf{x}) = (\sigma(\mathbf{x}), \textbf{c}(\mathbf{x},  \theta_r, \phi_r)),
\end{align}
where \( \mathbf{d} = (\theta_r, \phi_r) \) is the direction of the ultrasonic wave passing through point \( \mathbf{x} \) in the ultrasound volume $V$.

The synthesized new view images \( \{I_{new}^j\}_{j=1}^M \) enable clinicians to examine the target anatomy from multiple angles. AIA-UltraNeRF subsequently performs ultrasound localization by retrieving an initial plane \( P_{init} \) through offline image matching, which is optimized by minimizing a retrieval loss, as described in~\ref{chapter 4-3}. This initial plane serves as a coarse estimate of the standard plane, which refers to clinically validated imaging planes that systematically capture diagnostically critical anatomical structures.
\begin{align}
P_{init} = \arg\min_{P} \mathcal{L}_{retrieval}(P, \hat{P}),
\end{align}
where \( P \) represents the pose of the retrieved image, and \( \hat{P} \) indicates the pose of the standard plane. Refinement adjustments can utilize iNeRF, an iterative pose optimization method that estimates the 6-DoF pose by minimizing the difference between rendered and observed images, ultimately refining the pose to \( P_{final} \) described in~\ref{chapter 6-1}:
\begin{align}
P_{final} = P_{init} + \Delta P,
\end{align}
where $\Delta P$ is the fine-tuning offset.

\begin{figure}[!t]
\centering
\includegraphics[width=3.5in]{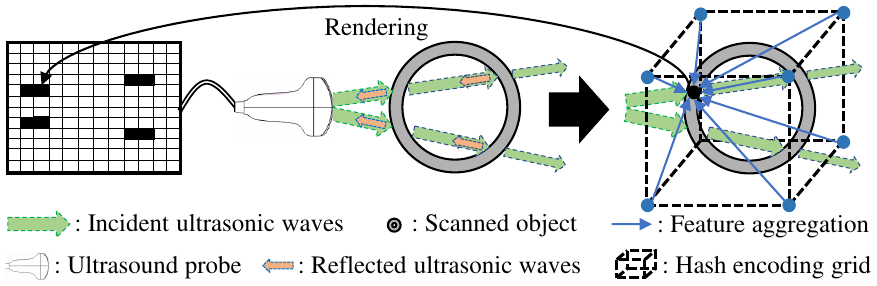}
\caption{Ultrasonic waves are reflected by tissue with differing acoustic impedance to form ultrasound images, and ultrasound images are rendered by aggregating acoustic impedance features.}
\label{fig3}
\end{figure}

\subsection{Ultrasound Reconstruction Method}\label{chapter 4-2}
\subsubsection{Ultrasound imaging principles and acoustic impedance}
Ultrasound imaging is a non-invasive, wide-ranging modality for real-time visualization of anatomical structures. It operates by emitting ultrasonic waves through a probe, which propagate through tissues and are partially reflected at boundaries with differing properties. These echoes are received and converted into electrical signals, which are then processed into B-mode images, as shown in Fig.~\ref{fig3}. A key physical principle underlying this process is acoustic impedance, defined as the ratio of sound pressure $p$ to particle velocity $u$, or the product of tissue density $\rho$ and the speed of sound $c$ ($\mathscr{Z} = p/u = \rho c$) \cite{ito2017acoustic}. Impedance differences between tissues result in the partial reflection and transmission of ultrasound waves at their boundaries. Since each type of tissue exhibits a different impedance, the strength of the reflected signal depends on the contrast between adjacent tissues. Consequently, the brightness of the B-mode image indirectly reflects the local acoustic impedance. We encode acoustic impedance using hash grids for ultrasound reconstruction to better capture this physical process. 

\subsubsection{Acoustic impedance feature characterization} 
The acoustic impedance at a given position $\mathbf{x}$ can be represented as an implicit function $Z_\Omega: \mathbf{x}\mapsto \mathscr{Z}$, where $\mathscr{Z}$ represents the acoustic impedance at $\mathbf{x}$. Assuming that a feature grid encompasses the point, as shown in Fig.~\ref{fig3}, the acoustic impedance at that point is modeled as an implicit function: 
\begin{align}
Z_\Omega: \mathcal{R}(\mathbf{x}) \mapsto\mathscr{Z} \quad \text{if} \quad \mathbf{x} \in V,
\end{align}
where $\mathcal{R}(\mathbf{x})$ is the representation of $\mathbf{x}$, which is defined as: $\mathcal{R}(\mathbf{x}) =  \chi ( \tilde{\mathcal{R}}(\mathbf{x}_1^*), \ldots, \tilde{\mathcal{R}}(\mathbf{x}_8^*)) $. $\mathbf{x}_1^*, \ldots, \mathbf{x}_8^*\in \mathbb{R}^3$ are the eight vertices wraps point $\mathbf{x}$ and \(\tilde{\mathcal{R}}(\mathbf{x}_i^*) \in \mathbb{R}^f\) is the feature vector at vertex \(\mathbf{x}_i^*\), where \(f\) is the feature dimension. Additionally, $\chi(\cdot)$ refers to trilinear interpolation for aggregating vertex features.

\subsubsection{Hash grid encoding}
A single grid with limited feature dimensions struggles to accurately represent the acoustic impedance at each point using trilinear interpolation. To overcome this limitation, multi-resolution grids are employed, dividing the space into finer resolutions. This approach allows for a more precise representation of varying acoustic impedances, thereby enhancing performance in ultrasound reconstruction.

Let $V_l \in \mathbb{R}^{N_l \times N_l \times N_l \times f}$ represent the grid of parameters at the $l$-th level, $l=1,\cdots, L$, where $L$ is the number of levels. $N_l$ denotes the size of the $l$-th grid in each dimension, with each vertex of the grid conceptually containing a feature vector of length $f$. However, if the resolution is increased in this manner, memory consumption will increase cubically with the rise in grid resolution. To address this issue, the dense 3D array $V_l$ is replaced by a hash table $V_l^{\text{hash}} \in \mathbb{R}^{N_h \times f}$, where $N_h$ is the size of the hash table and $N_h \ll (N_l)^3$. Therefore, the queried acoustic impedance at point $\mathbf{x}$ of the grid $V^{hash}_l$ is summarized as follows, as shown in Fig.~\ref{fig2} (b):
\begin{enumerate}
    \item The input $\mathbf{x}\in \mathbb{R}^3$ is scaled by the $l$-level grid to extract the index $(i, j, k)$ of the grid $V^{hash}_l$. 
    \item Mapping the index $(i, j, k)$ to a hash code with a spatial hash function. $V^{hash}_l(i, j, k) = V_l^{\text{hash}}(i \oplus j \pi_1 \oplus k \pi_2 \mod N_h)$, where $\pi_1$ and $\pi_2$ are two large primes, and $\oplus$ denotes the bit-wise XOR operation.
    \item Encoding $V^{hash}_l(i, j, k)$ to feature dimension $f$. Finally, the acoustic impedance of the point is obtained through trilinear interpolation $\chi(\cdot)$.
\end{enumerate}

Acoustic impedance can be efficiently aggregated using a hash table structure across multiple levels, which significantly reduces memory usage and enhances computational efficiency. Encoded features from different levels will be concatenated $V^{hash}(\mathbf{x}) =[V^{hash}_1(\mathbf{x}), \cdots, V^{hash}_L (\mathbf{x})]$ and transmitted $V^{hash}(\mathbf{x})$ to the decoder $F_{\Theta}$ for color rendering.

\subsubsection{Color rendering}

The pixel color of the selected point $\mathbf{x}$ is decoded using a tiny MLP that takes both hashed features and wave direction as inputs, as the acoustic impedance at that point is represented across hash grids.
The pixel color can ultimately be obtained through MLP decoding:
\begin{align}
I^{'}(\mathbf{x}, \mathbf{d}) = F_{\Theta}(V^{hash}(\mathbf{x}), \mathbf{Y}(\theta_r, \phi_r)),
\label{eq6}
\end{align}
where $\Theta$ denotes the parameters of the MLP $F_{\Theta}$, and $\mathbf{Y}(\theta_r, \phi_r)$ is the spherical harmonic of the direction $\mathbf{d}$.

\subsection{Ultrasound Localization Method}\label{chapter 4-3}
\subsubsection{Self-distilled hashing}
Due to high noise levels and blurred boundaries in ultrasound images, it is challenging for a single neural network to extract precise features. The robustness of the encoder model can be improved by utilizing two separate augmentation groups: weakly transformed $\mathcal{T}_t$ for the teacher model and strongly transformed $\mathcal{T}_s$ for the student model. This approach enables the neural network to accurately extract features from ultrasound images in high-noise environments while enhancing the model's consistency in feature extraction, thus preventing similar inputs from producing varying outputs. 

For a given ultrasound image $\hat{I}$, various transformations are applied to the image before it is input into the model as $\hat{u}_t = \mathcal{T}_t(\hat{I})$ and $\hat{u}_s = \mathcal{T}_s(\hat{I})$. The deep hash network takes $\hat{u}_t$ and $\hat{u}_s$ as inputs and produces the corresponding hash codes $h_t$ and $h_s$. Subsequently, the self-distilled hashing loss is computed:
\begin{align}
\mathcal{L}_{dhl}({h}_t, {h}_s) = 1 - \mathcal{S}({h}_t, {h}_s),
\end{align}
where \( \mathcal{S}(h_t, h_s) \) denotes the cosine similarity between the output hash codes of the teacher and the student.

Optimizing the feature extraction network with $\mathcal{L}_{dhl}$ results in the alignment of $h_t$ and $h_s$, reducing the representation discrepancy between the two different output binary codes.

\subsubsection{Dual supervision and Gaussian estimator}

Additional supervised signals are necessary, alongside self-distilled hashing, to effectively learn distinct categories and capture similar features within the same category. Both the teacher and student hash codes are used to compute the loss, which helps prevent the student model from collapsing under strong transformations.

A collection of trainable hash proxies ${p}_\epsilon$, similar to fully connected layers, is employed to assess the relationship between the encoded hash value $h$ and the image label $y$. The class-wise prediction $\mathcal{P}$ with $h$ is calculated as follows:
\begin{align}
\mathcal{P}_t = \left[ \mathcal{S}(p_{\epsilon_1}, {h}_t), \mathcal{S}(p_{\epsilon_2}, {h}_t), \ldots, \mathcal{S}(p_{\epsilon_{N_{cls}}}, {h}_t) \right], \\
\mathcal{P}_s = \left[ \mathcal{S}(p_{\epsilon_1}, {h}_s), \mathcal{S}(p_{\epsilon_2}, {h}_s), \ldots, \mathcal{S}(p_{\epsilon_{N_{cls}}}, {h}_s) \right],
\end{align}
where $p_{\epsilon_i}$ is a hash proxy assigned to each of the $i$-th class and $N_{cls}$ denotes the number of classes to be distinguished. Subsequently, we use $\mathcal{P}$ to learn the similarity with the class label $y$ through cross-entropy:
\begin{align}
\mathcal{L}_{hpl} &= \mathcal{L}(y, \mathcal{P}_{t}, \tau_t) +\mathcal{L}(y, \mathcal{P}_{s}, \tau_s) =\nonumber\\  
& \mathcal{H}(y, \text{Softmax}(\mathcal{P}_t/\tau_t)) + \mathcal{H}(y, \text{Softmax}(\mathcal{P}_s/\tau_s)),
\end{align}
where $\tau$ is a hyperparameter for temperature scaling, and $\mathcal{H}(u,v)=-\sum_ku_k\text{log}v_k$ denotes the cross-entropy.

Since the final learned hash code is binary, the hash code generated after applying the sign function will be more accurate when the learned real value $h$ is closer to $\pm 1$. A predefined Gaussian distribution quantization estimator $g(h)$ with a mean of $r$ ($r=\pm 1$ for $g^{\pm}$) and a standard deviation of $\varrho$ as $g(h) = \exp ( -\frac{(h - r)^2}{2\varrho^2} )$ is applied to evaluate the binary likelihood of the learned real value, forming the binarization estimation loss $\mathcal{L}_{ql}({h}_t,{h}_s)$. For more details, please refer to \cite{jang2022deep}.

\subsubsection{Similarity valuation}

Since subtle changes in probe position can lead to significant differences in ultrasound images, each image must be treated as an independent class. Localization, therefore, focuses on extracting the probe pose corresponding to a given ultrasound image. Directly using the Hamming distance to measure the hash distance between two ultrasound images always yields a value of $1$. However, the differences in the corresponding probe poses are not constant. This highlights the need for a customized similarity evaluation method to capture the relationship between images and their corresponding poses.

To address this issue, we define a similarity calculation method based on label analysis. Given two labels $y_1$ and $y_2$, the first step is to identify the positions of the non-zero elements in each label:
\begin{align}
\text{pos}_1 &=  \{ i \mid y_1[i] \neq 0 \},\nonumber \\
\text{pos}_2 &=  \{ j \mid y_2[j] \neq 0 \}.
\end{align}

Next, we compute the similarity score based on the positional difference: 
\begin{align}
\mathfrak{S} = 
\begin{cases} 
\frac{1}{|\text{pos}_1 - \text{pos}_2|} & \text{if} \quad i \neq j, \\
\quad 1 & \text{if}\quad i = j.
\end{cases}
\end{align}

This similarity evaluation aids in identifying the image that is most similar to a given query. Once the most similar image is retrieved, the pose associated with that image is extracted. To quantify the accuracy of this retrieval process, we calculate the retrieval error using the following formula: 
\begin{align}
\mathcal{E} =  \cos^{-1}\left(\frac{\text{trace}(R_q^T R_r) - 1}{2}\right),
\label{eq13}
\end{align}
where $
R_{q,r} = R_z(\vartheta_z,\varphi_z) R_y(\vartheta_y,\varphi_y) R_x(\vartheta_x,\varphi_x)
$, $(\vartheta_x,\vartheta_y,\vartheta_z)$ and $(\varphi_x,\varphi_y,\varphi_z)$ represent the roll, pitch, and yaw of the probe pose corresponding to the query and retrieved images, respectively. It is important to note that the probe poses are relative to the world coordinate frame. 

\subsection{Loss Function}\label{chapter 7}

Each epoch randomly selects an image along with its corresponding pose for reconstruction. The acoustic impedance of the tissue at each pixel is aggregated using hash grids, while the color is rendered using spherical harmonics, as shown in Eq.~(\ref{eq6}). The loss is computed against the ground truth color to update the network as follows:
\begin{align}
\mathcal{L}_{recon} = \frac{1}{N_{rec}}\sum_{i=1}^{N_{rec}} \| I^{'}(\mathbf{x}, \mathbf{d}) - I(\mathbf{x}, \mathbf{d}) \|_2^2,
\end{align}
where ${N_{rec}}$ represents the number of selected pixels, $I(\mathbf{x}, \mathbf{d})$ signifies the ground truth color of the pixel, and $I^{'}(\mathbf{x}, \mathbf{d})$ denotes the rendered color of the pixel.

For localization, a training mini-batch of ultrasound images is provided, with each image $\hat{I}_i$ assigned a corresponding label $y_i\in \{0, 1\}^{N_{cls}}$. We input images with various transformations into both the teacher and student networks to generate real-valued $h$. Subsequently, we apply the loss derived from self-distilled hashing, dual supervision, and Gaussian distribution estimation to optimize the network:
\begin{align}
\mathcal{L}_{local} = \sum_{i=1}^{N_{loc}}(\mathcal{L}_{hpl}+\lambda_1\mathcal{L}_{dhl}+\lambda_2\mathcal{L}_{ql}),
\end{align}
where ${N_{loc}}$ denotes the mini-batch, and $\lambda_1$ and $\lambda_2$ are hyperparameters.

\begin{figure}[!t]
\centering
\includegraphics[width=3.5in]{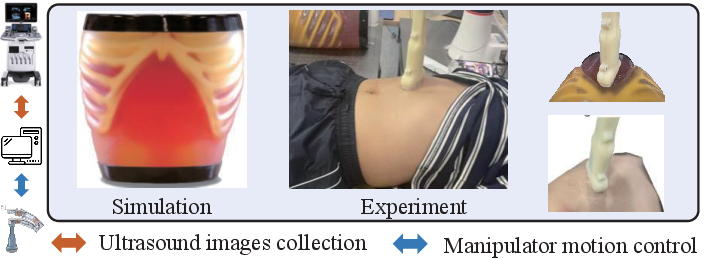}
\caption{Experimental setup: Simulation is conducted using the ABDFAN phantom, and the experiment involves human subjects.}
\label{fig4}
\end{figure}

\begin{figure}[!t]
\centering
\includegraphics[width=3.5in]{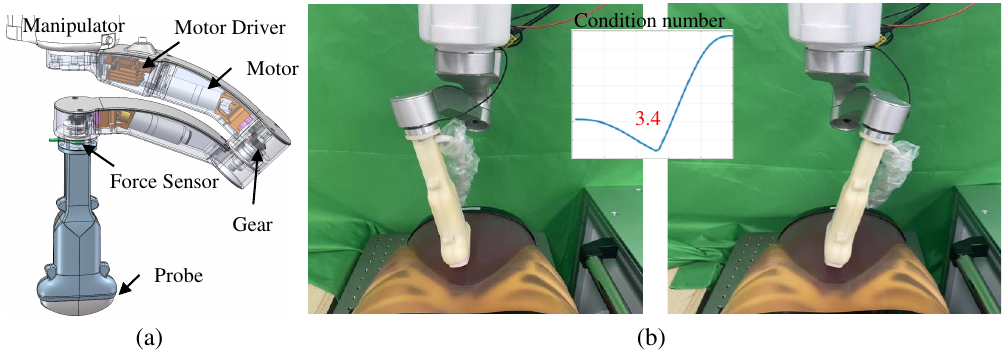}
\caption{(a) Spherical series mechanism for ultrasound scanning. (b) Motion flexibility of the spherical RCM mechanism during scanning.}
\label{fig1-1}
\end{figure}



\section{Experimental Results}\label{chapter 8}



\begin{figure}[!t]
\centering
\includegraphics[width=3.5in]{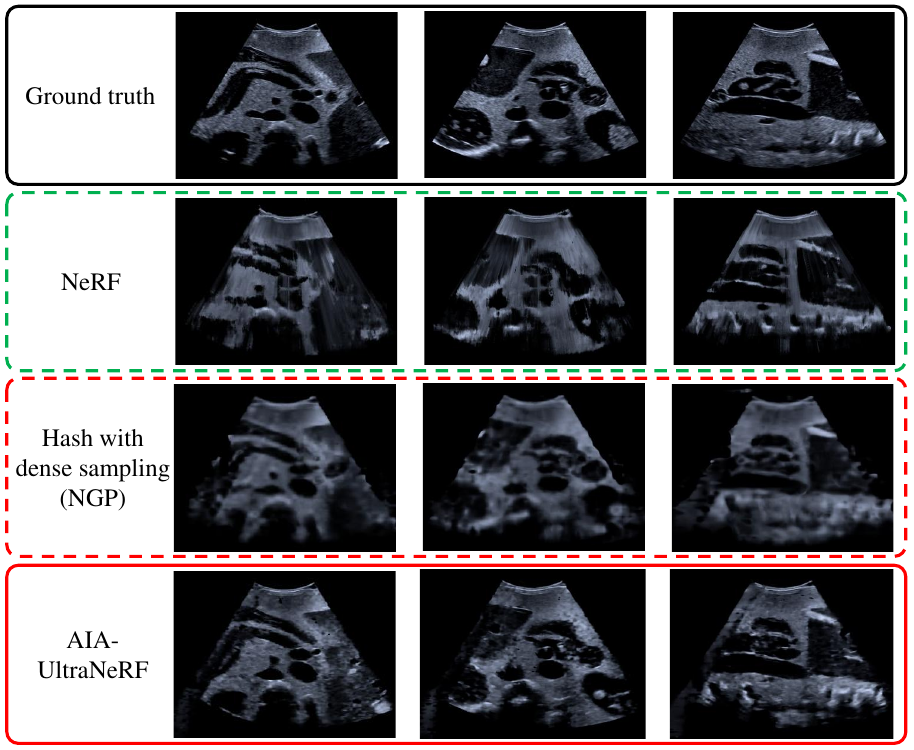}
\caption{Qualitative comparison. New view rendering results of reconstructed ultrasound volume on the phantom in the test set. 
}
\label{fig6}
\end{figure}

\begin{figure}[!t]
\centering
\includegraphics[width=3.5in]{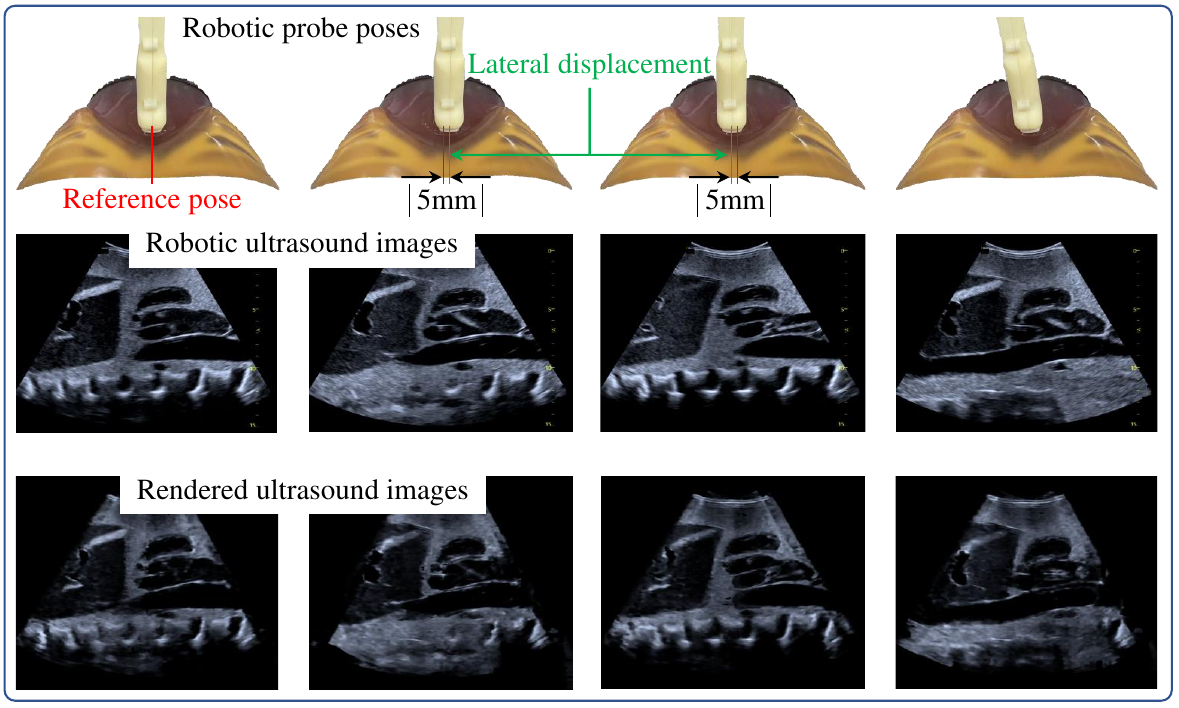}
\caption{Qualitative comparison of entirely new view images rendering beyond the normal acquisition dataset with those acquired by RUSS. The first row shows the robotic probe poses, the second row presents the ultrasound images acquired by RUSS at those poses, and the third row shows the corresponding images rendered by AIA-UltraNeRF with the same probe poses.
}
\label{fig7}
\end{figure}

To validate the effectiveness of the proposed AIA-UltraNeRF for reconstructing ultrasound maps and localizing from the reconstructed map, we employed an ABDFAN ultrasound examination phantom (Kyoto Kagaku, Japan). Additionally, we conducted experimental verification on human subjects, collecting liver and kidney ultrasound images from ten male human subjects for reconstruction. We then performed ultrasound localization using the reconstructed ultrasound maps of the human subjects. Informed consent was obtained from all volunteers. Ethical approval for this study was granted by the ethical review committee of Hefei University of Technology (Grant No. HFUT20241204001H). The overall experimental setup is shown in Fig.~\ref{fig4}. Next, we demonstrate the superiority of the proposed AIA-UltraNeRF through phantom simulations and experiments involving human subjects.

In the reconstruction module, $L=16$, $f=8$, and $N_h=2^{21}$ are selected for the hash grids. An MLP with two layers and 128 neurons is utilized to predict density. The output is processed using a ReLU activation function to ensure that the values remain non-negative. Additionally, a 15-dimensional feature vector is created, which is combined with ultrasound direction encoded using spherical harmonic encoding. This combined input is fed into another MLP, also consisting of two layers and 128 neurons, and is processed with ReLU activation to output the corresponding color. Different backbones, such as ResNet50 from torchvision, ViT-base, and DeiT-base from timm, are employed as image feature encoders  \cite{he2016deep, dosovitskiy2020image, touvron2021training} for the localization module of AIA-UltraNeRF. We train our model with 50,000 iterations for reconstruction and 400 iterations for localization. We use the Aadam optimizer for the reconstruction module and the standard Adam optimizer for the localization module. All experiments were conducted on a workstation equipped with an Intel Xeon Silver 4210R CPU and an NVIDIA GeForce RTX 3090 GPU. 

For phantom data acquisition, the ultrasound probe was rotated along a circular trajectory with a diameter of 2 cm, capturing 72 images: 58 images were used for training, and 7 images each were used for validation and testing. For human data acquisition, we adopted a fixed-point rotational scanning approach, where the ultrasound probe remained in a fixed position and rotated around its central axis. The scanning was performed with a step size of 5° per frame, capturing 37 images per individual: 30 images were used for training, 3 images for validation, and 4 images for testing. The resolution of the images is 1024$\times$768 (0.23 mm/pixel). The ground truth pose of each ultrasound image in the reconstruction module was obtained using forward kinematics based on the joint angles of RUSS, and the transformation between the ultrasound probe and the robot flange was calibrated prior to image acquisition. In the intra-subject localization module, the image pose refers to its rendered poses, while in the inter-subject localization module, the image pose corresponds to its rendered poses in other subjects.

To comprehensively evaluate the performance of AIA-UltraNeRF, we adopt several widely used metrics to measure reconstruction quality, rendering efficiency, and pose estimation accuracy. PSNR (peak signal-to-noise ratio) measures reconstruction fidelity, with higher values indicating better quality. SSIM (structural similarity index measure) evaluates the structural similarity between images, where values closer to 1 indicate higher similarity \cite{wang2004image}. LPIPS (learned perceptual image patch similarity, with embeddings extracted from the pre-trained AlexNet) quantifies perceptual distance using deep features, with lower scores representing better similarity \cite{zhang2018unreasonable}. Inference time refers to the duration required to render a complete ultrasound image, indicating computational efficiency. Error, computed using Eq.~(\ref{eq13}), measures the angular error between the predicted and ground truth images, with lower values indicating higher accuracy.

\subsection{Ultrasound Imaging and Localization System}\label{chapter 5}

Separating scanning and diagnostic workflows necessitates a flexible scanning mechanism that offers a large workspace for image acquisition and precise ultrasound localization. Transitioning from one tilt angle to its symmetrical counterpart requires a significant range of motion for a robotic arm, thereby increasing the demand for movement space. Furthermore, to prevent collisions between the manipulator and the patient's body or surrounding equipment, the trajectory of the manipulator's movement and its initial position must be meticulously planned. Consequently, designing a flexible mechanism for the URIPL in RUSS is essential to facilitate continuous and adaptable movement within a constrained space while ensuring high image quality and patient safety.

To address these requirements, we designed a RUSS with a customized spherical RCM mechanism, as shown in Fig.~\ref{fig1-1}(a). The complete RUSS primarily consists of four components: a 7-DoF robotic manipulator (xMate 3 Pro, ROKAE, China), a customized RCM mechanism, a convex array probe (S1-8C, VINNO, China), and an ultrasound machine (N75, VINNO, China). The remote center will be positioned below the superficial surface of the patient's skin. During the scanning process, the remote center will remain stationary to ensure the patient's safety, similar to a minimally invasive surgery system \cite{lum2006optimization}. This spherical series mechanism can achieve fixed-point scanning in any direction and adapt to the requirements of various tasks. 


The spherical RCM mechanism occupies less space. 
The condition number presented in Fig.~\ref{fig1-1}(b) reflects the manipulability of the spherical RCM mechanism. A condition number of approximately 3.4 indicates stable and well-conditioned motion during scanning, thereby avoiding singularities and ensuring flexible probe control. The spherical RCM mechanism demonstrates exceptional operational flexibility and stability within a confined space, making it ideal for precise abdominal scanning. Furthermore, the movement space provided by the RCM mechanism is significantly larger than that in \cite{zhang2024development}.

\subsection{Baselines}\label{chapter 6-0}
To evaluate the effectiveness of AIA-UltraNeRF, we compare it to two representative NeRF-based reconstruction methods. NeRF is a widely used implicit 3D representation that employs an MLP to predict colour and density at any given spatial coordinate and viewing direction \cite{mildenhall2021nerf}, as applied by Ultra-Malin for ultrasound reconstruction \cite{zhang2025ultra}. Instant-NGP, implemented purely in PyTorch, models a scene by leveraging multi-resolution hash encoding and a compact MLP \cite{mueller2022instant}. This method significantly reduces training time while maintaining high rendering quality, establishing it as a robust baseline for 3D reconstruction tasks.

\begin{figure}[!t]
\centering
\includegraphics[width=3.5in]{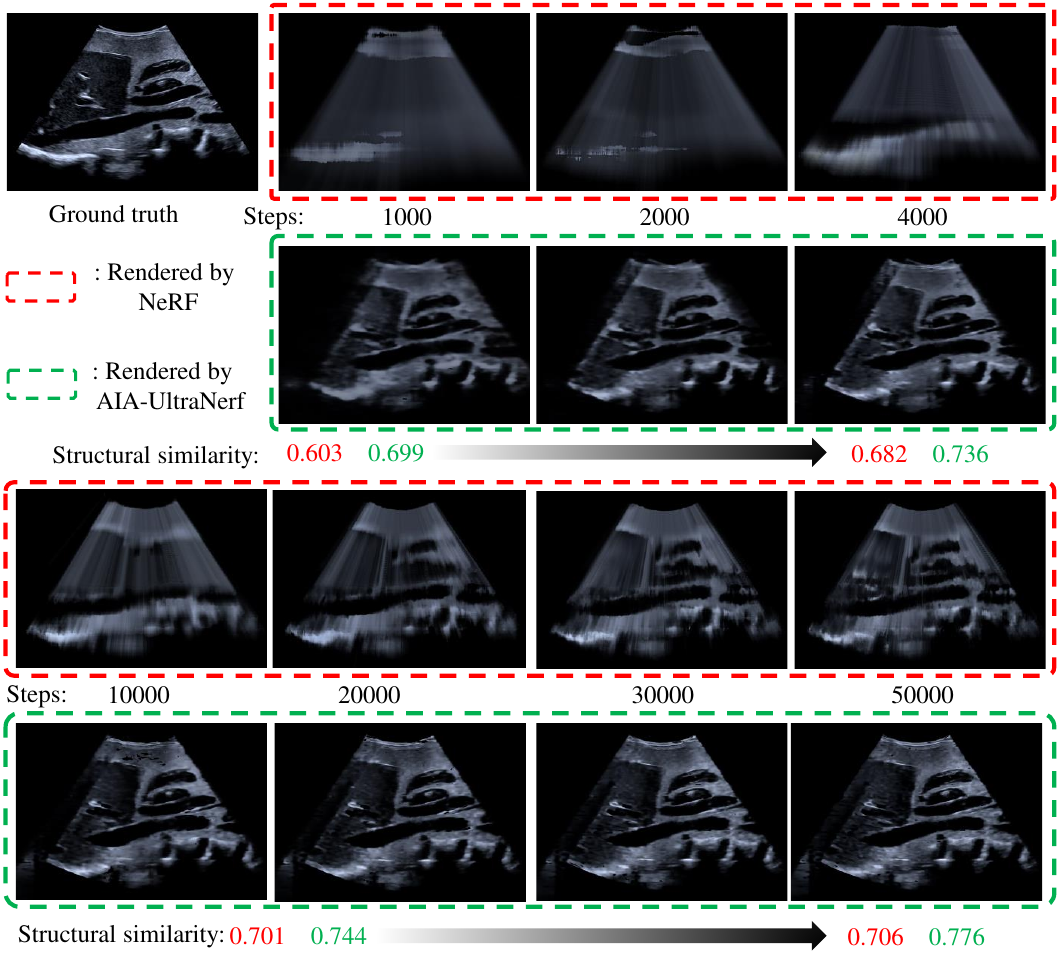}
\caption{Visual comparison of rendering results from NeRF and AIA-UltraNeRF at different training steps with the same number of pixels sampled at each step. The top two rows show results from NeRF (red dashed box), and the bottom two rows show results from AIA-UltraNeRF (green dashed box). Ground truth is shown at the top left.}
\label{fig8}
\end{figure}

\begin{figure}[!t]
\centering
\includegraphics[width=3.5in]{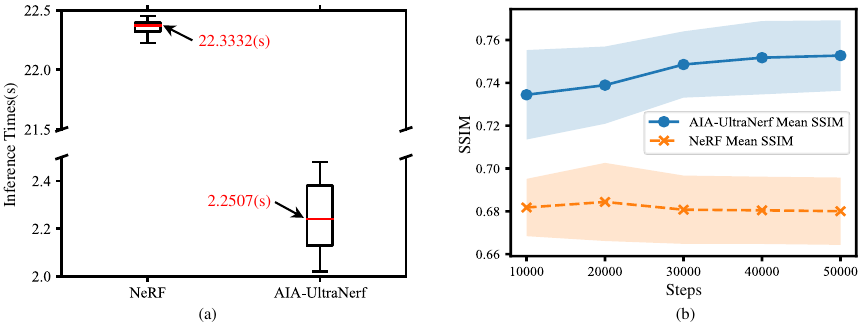}
\caption{(a) Inference time comparison: AIA-UltraNeRF renders ultrasound images 9.9× faster than NeRF. (b) Reconstruction efficiency comparison: AIA-UltraNeRF achieves consistently higher SSIM than NeRF with the same number of pixels sampled at each step.}
\label{fig9-1}
\end{figure}

\subsection{Simulation on Phantom}\label{chapter 6-1}

The rendering results of the normal annular scanning using different methods are shown in Fig.~\ref{fig6}. These results indicate that AIA-UltraNeRF exhibits remarkable performance compared to other methods employing various sampling strategies. NeRF relies on multi-view data to approximate the density distribution; however, inconsistencies can lead to reconstruction errors. Its dependence solely on learning acoustic impedance often results in unreliable outputs, including completely black images. NGP involves learning the acoustic impedance of various tissues through dense sampling, complicating detailed differentiation and leading to blurred reconstructions. In contrast, AIA-UltraNeRF demonstrates superior performance in acoustic impedance learning, effectively capturing the acoustic impedance at each point. The quantitative analysis results are displayed in Tab.~\ref{Tab2}. AIA-UltraNeRF achieves a PSNR of 20.2560, outperforming NeRF by 2.57 dB and NGP by 0.14 dB, indicating more accurate ultrasound reconstruction. AIA-UltraNeRF obtains an SSIM of 0.7569, which is 11.5\% higher than NeRF and 4.99\% higher than NGP.
Additionally, AIA-UltraNeRF achieves a significantly lower LPIPS (AlexNet) score of 0.2980, representing a 30.2\% reduction compared to NeRF and a 26.4\% reduction compared to NGP. 

\begin{table}
\caption{\textbf{Quantitative comparison. New view rendering results on the phantom in the test set using different methods.}}
\centering
\setlength{\tabcolsep}{4mm}{\begin{tabular}{cccc}
\toprule
   Method    &   PSNR($\uparrow$)  &  SSIM($\uparrow$)    &  LPIPS($\downarrow$) \\
\midrule
NeRF & 17.6842 & 0.6788 & 0.4271 \\
NGP    & 20.1199 & 0.7213 & 0.4050 \\
AIA-UltraNeRF   & \textbf{20.2560} & \textbf{0.7569} & \textbf{0.2980} \\
\bottomrule
\end{tabular}}
\label{Tab2}
\end{table}

The rendering results of entirely new-view ultrasound images, beyond the normal acquisition dataset, are presented in Fig.~\ref{fig7}. The ultrasound images acquired by RUSS with the same probe pose are compared to demonstrate the accuracy of the rendered new-view images. The reconstruction utilizes images collected from a normal circular scanning method, while Fig.~\ref{fig7} illustrates that ultrasound images can be rendered not only at the center of the circle but also at offset positions and various tilt angles. Furthermore, the broader coverage provided by the circular scan facilitates high-quality rendering of tilted scan images within the region. AIA-UltraNeRF offers valuable support for exploring unscanned areas with minimal view scans and for rendering images in regions without predetermined probe placement.

Fig.~\ref{fig8} shows a rendering example of NeRF with dense sampling and AIA-UltraNeRF at different training steps, highlighting improvements in SSIM alongside reconstruction efficiency. The red dashed boxes contain rendered images of NeRF, while the green dashed boxes contain rendered images of AIA-UltraNeRF. The results indicate that NeRF yields a lower SSIM at the same training step. These results underscore the advantages of AIA-UltraNeRF in terms of efficiency and effectiveness, demonstrating that it can achieve an improvement of 11.5\% in SSIM, which surpasses \cite{wysocki2024ultra, yeung2021implicitvol}. It is noteworthy that to render 20 images, as shown in Fig.~\ref{fig9-1} (a), NeRF takes approximately 22 seconds to render a single image, while AIA-UltraNeRF achieves an inference time of around 2 seconds, making it nearly 9.9× faster. Meanwhile, Fig.~\ref{fig9-1} (b) presents the results of three training runs, indicating that AIA-UltraNeRF achieves a significantly higher SSIM than NeRF at the same step count, with both models utilizing the same number of rays per epoch, thereby demonstrating enhanced reconstruction efficiency compared to NeRF.


\begin{figure}[!t]
\centering
\includegraphics[width=3.5in]{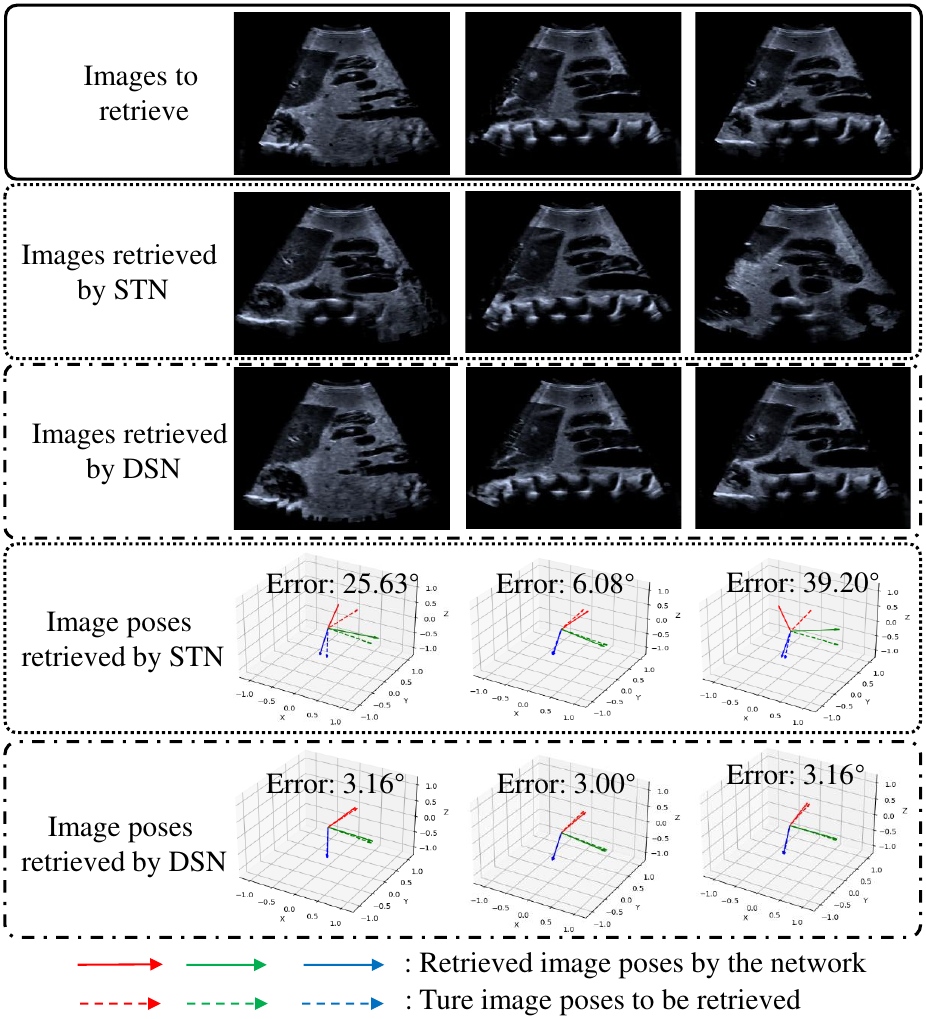}
\caption{Qualitative comparison of ultrasound localization on the phantom, and their corresponding probe poses. The second and third rows are the retrieval results of STN and DSN. The red, green, and blue arrows in the last two rows correspond to the image pose's $x,y, z$ axes with angular error.}
\label{fig11}
\end{figure}

\begin{figure}[!t]
\centering
\includegraphics[width=3.5in]{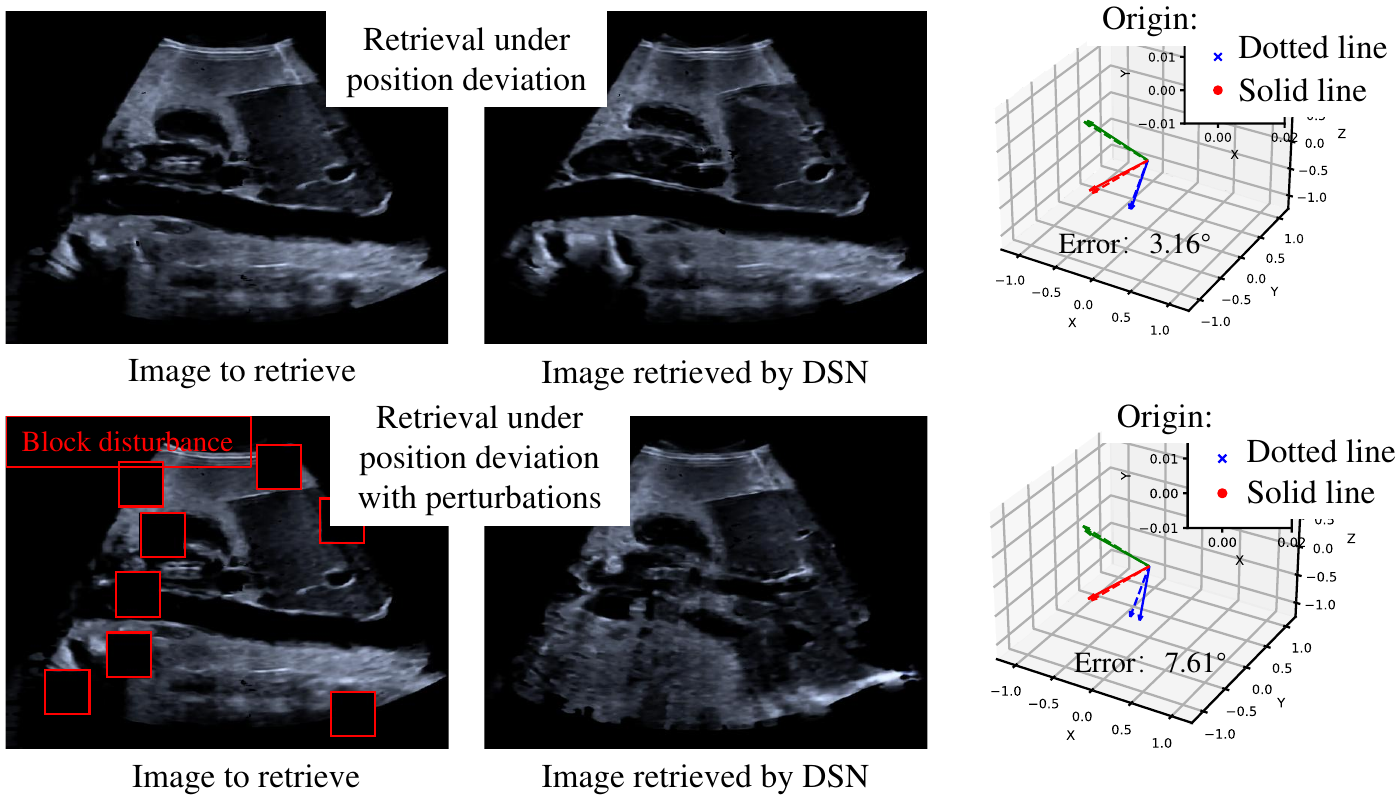}
\caption{Ultrasound localization results under position deviation and position deviation with black block disturbance. The top row shows the localization results under position deviation, and the bottom row shows the localization results under position deviation with disturbance. Next, from left to right, are the images to be retrieved, the retrieved images, and the retrieval errors.}
\label{fig11-1}
\end{figure}

\begin{figure}[!t]
\centering
\includegraphics[width=3.2in]{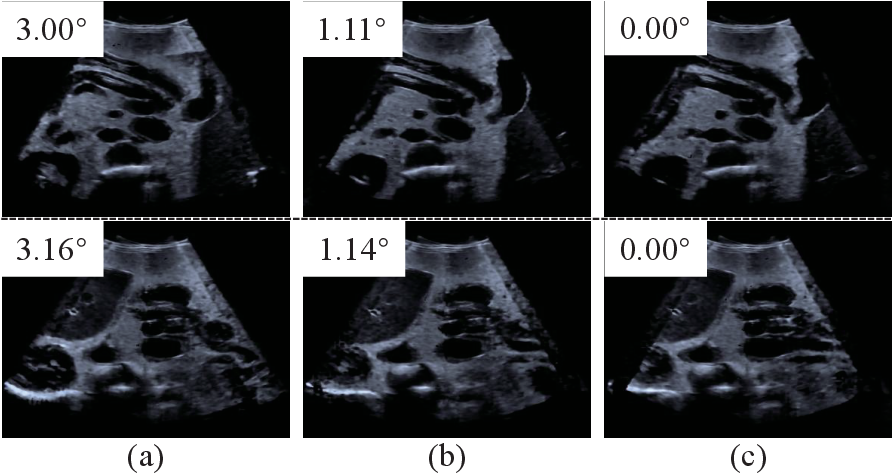}
\caption{(a) Images of the initial pose $P_{init}$ obtained from the offline localization given by AIA-UltraNeRF. (b) The refined results after applying the iNeRF-based pose online refinement. (c) The ground truth. The value in the upper left corner represents the angular difference from the ground truth.}
\label{fig11-2}
\end{figure}

\begin{figure}[!t]
\centering
\includegraphics[width=3.5in]{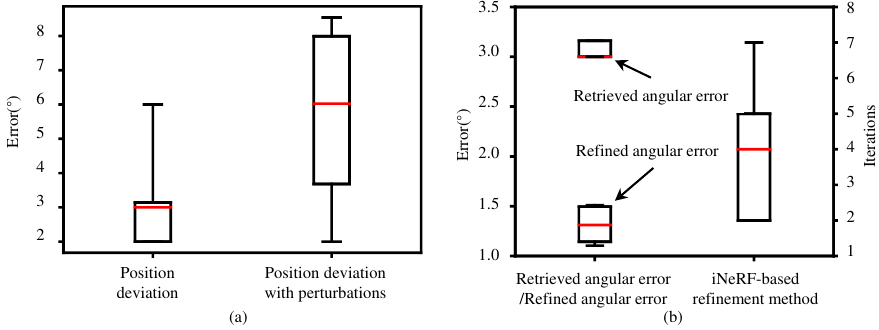}
\caption{(a) Retrieval results under position deviation and position deviation with perturbations. (b) Comparison of pose estimation retrieved from AIA-UltraNeRF with online refinement using iNeRF, showing angular error before and after refinement on the left, and iteration count on the right.}
\label{fig11-3}
\end{figure}

\begin{table}
\caption{\textbf{Quantitative comparison of localization errors on the phantom with different backbones.}}
\centering
\setlength{\tabcolsep}{2.5mm}{\begin{tabular}{cccc}
\toprule
  Backbone     &   DSN(ours)  &  STN    & TSN(\cite{shao2021deep, jang2022deep}) \\
\midrule
ResNet & \textbf{3.70°(100\%)} & 3.59°(70\%) & 3.64°(75\%) \\
ViT    & 3.07°(75\%) & 3.85°(55\%) & 2.94°(75\%) \\
DeiT   & 4.03°(75\%) & 2.87°(70\%) & 2.88°(75\%) \\
\bottomrule
\end{tabular}}
\label{Tab3}
\end{table}

In the retrieval-based localization simulations, we rendered 900 ultrasound images from the reconstructed 3D ultrasound map, designating ninety images for verification. These images were rendered with a tilt range from 0° to 15° at 3° rotation intervals at a fixed point in the annular scan. The query images were not included in the training dataset to evaluate the network's ability to localize within new ultrasound scans. We focus on comparing the performance of the dual-supervised network (DSN) with the single-teacher network (STN) and the teacher-student network without student supervision (TSN). The final localization error for 20 randomly selected images, calculated using Eq.~(\ref{eq13}), along with the success ratios for retrieval accuracy of less than 10°, is presented in Tab.~\ref{Tab3}. AIA-UltraNeRF with a ResNet backbone achieves a localization error of $3.70^{\circ}$ and a success rate of 100\%, indicating robust and accurate retrieval performance. Compared to DSN with ViT ($3.07^{\circ}$, 75\%) and DeiT ($4.03^{\circ}$, 75\%), ResNet achieves a lower angular error than DeiT while exhibiting significantly higher reliability compared to both transformer-based backbones. STN shows slightly lower localization errors for ViT and DeiT, but its success rates drop to 55\% and 70\%, respectively, which are substantially lower than those of DSN with ResNet. Similarly, TSN yields comparable localization errors (e.g., $3.64^{\circ}$ for ResNet and $2.94^{\circ}$ for ViT), but the success rate remains fixed at 75\% across the backbones. These results quantitatively demonstrate that AIA-UltraNeRF with the ResNet backbone offers the most accurate localization, highlighting its effectiveness in ultrasound localization even under tilted probe poses.
Fig.~\ref{fig11} illustrates a comparison of various retrieval methods for ultrasound localization, showcasing the query images, the images retrieved using STN and DSN, and their corresponding probe poses. The retrieved probe positions (dashed arrows) are compared with the ground truth (solid arrows), where the errors indicate angular deviations. The results demonstrate that AIA-UltraNerf, which is trained using DSN, retrieves images that are more similar to the query, achieving significantly lower angular errors, with deviations as small as 1.00° compared to STN, which exhibits larger discrepancies of up to 39.20°.

\begin{figure}[!t]
\centering
\includegraphics[width=3.5in]{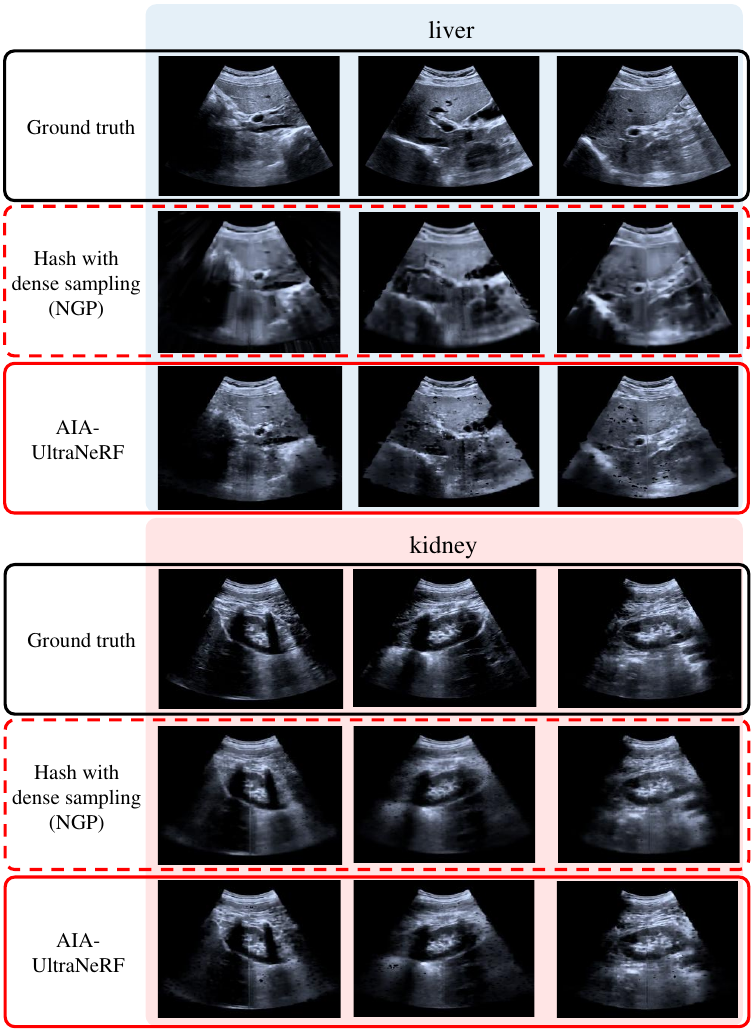}
\caption{Qualitative Comparison. New view rendering results of reconstructed ultrasound volumes on human subjects in the test set. 
}
\label{fig12}
\end{figure}

Note that even if the position of the image to be retrieved has some deviation within a circle with a diameter of 2 cm, AIA-UltraNeRF can still accurately determine its pose, as demonstrated in Fig.~\ref{fig11-1}. To further assess robustness, we randomly added black blocks to simulate perturbations, and AIA-UltraNeRF successfully provided effective localization. As shown in Fig.~\ref{fig11-3} (a), the average results of five experiments indicate that AIA-UltraNeRF maintains reliable localization. The left box plot of Fig.~\ref{fig11-3} (a) illustrates the error distribution under position deviation alone, while the right box plot reflects the increase in error when additional perturbations are introduced. These results demonstrate that AIA-UltraNeRF achieved the localization tasks with errors of less than 10° under challenging conditions. Furthermore, leveraging the offline initial pose provided by AIA-UltraNeRF, we applied iNeRF for online pose refinement to \( P_{final} \), as shown in Fig.~\ref{fig11-2}. We conducted five experiments, and the results presented in Fig.~\ref{fig11-3} (b) confirm an improvement in pose accuracy following online refinement. The left side of Fig.~\ref{fig11-3} (b) displays the refined angular error, while the right side indicates the number of iterations required by the Monte Carlo iNeRF-based refinement method. This highlights the significance of accurate initial pose selection provided by AIA-UltraNeRF in ultrasound localization using the Monte Carlo iNeRF-based online refinement approach.

\subsection{Experiment on Human subjects}

\begin{table}
\caption{\textbf{Average Quantitative comparison. New view rendering results on human subjects in the test set using different methods.}}
\centering
\setlength{\tabcolsep}{2.5mm}{\begin{tabular}{ccccc}
\toprule
   Method    &   PSNR($\uparrow$)  &  SSIM($\uparrow$)    &  LPIPS($\downarrow$) \\
\midrule
NGP    & 22.1358 & 0.7920 & 0.3154 \\
AIA-UltraNeRF
    & \textbf{22.7414} & \textbf{0.8168} & \textbf{0.2881} \\
\bottomrule
\end{tabular}}
\label{Tab4}
\end{table}

\begin{figure}[!t]
\centering
\includegraphics[width=3.5in]{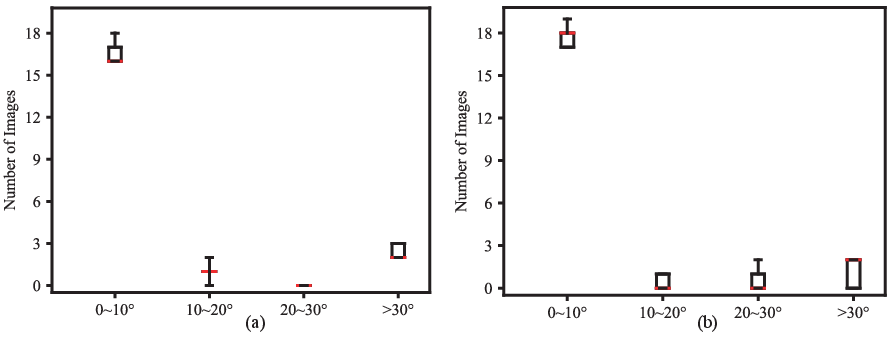}
\caption{(a) Statistical distribution of angular errors with 20 images in intra-subject on the liver using the DSN localization model. (b) Statistical distribution of angular errors with 20 images in intra-subject on the kidney using the DSN localization model.}
\label{fig13-1}
\end{figure}

\begin{figure}[!t]
\centering
\includegraphics[width=3.5in]{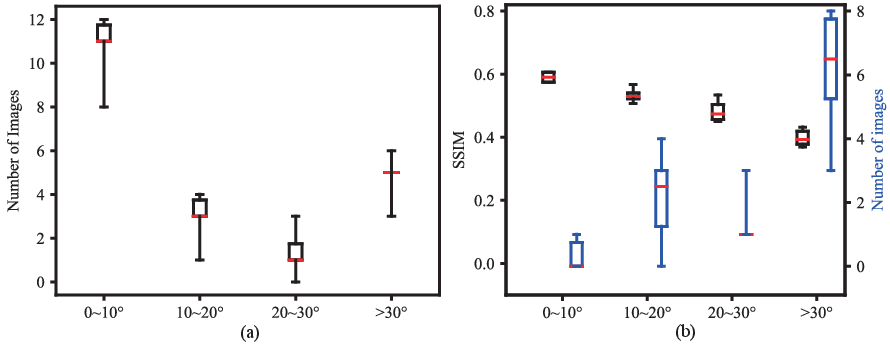}
\caption{(a) Statistical distribution of angular errors with 20 images in cross-subject across 6 different train-test volunteer combinations using the DSN localization model, where the model is trained on one volunteer and tested on the other two. (b) Statistical distributions of angular errors and the corresponding SSIMs across different ranges in inter-subject with 6 different train-test volunteer combinations. The DSN localization model is trained and tested on one volunteer, while the images for testing are retrieved from the other two volunteers. 
}
\label{fig13-2}
\end{figure}

We applied AIA-UltraNeRF to ten human subjects to further evaluate its effectiveness. Five volunteers were scanned to acquire liver data, and the remaining five were scanned to acquire kidney data. In human subjects, complex tissue interfaces generate intricate acoustic reflections, while physiological activities such as breathing and blood flow induce positional and morphological changes that destabilize ultrasound imaging and decrease clarity. As a result, ultrasound reconstruction and localization become increasingly challenging.
Fig.~\ref{fig12} displays the rendered new view images following ultrasound reconstruction on different volunteers of the liver and kidney. This demonstrates the validity of AIA-UltraNeRF in modeling the acoustic impedances of various tissues, thereby enabling new image synthesis for diagnostic purposes. 
It is evident that, in terms of clarity and accuracy, the color of the tissue can be effectively modeled by learning the acoustic impedance of tissues at various locations. However, NGP may introduce redundancy into the information and blur the image. The results of the average quantitative analysis are presented in Tab.~\ref{Tab4}. The fixed-point rotational scanning approach applied to human subjects reduces motion artefacts more effectively than the circular trajectory used on the phantom, resulting in improved reconstruction quality for human data.

We first conducted localization experiments using the DSN localization model in an intra-subject setting on both liver and kidney. The results, visualized in Fig.~\ref{fig13-1}, show the distribution of localization errors for intra-subject scenarios and various train-test volunteer combinations. Fig.~\ref{fig13-1} suggests that the DSN localization model can achieve low errors when trained and tested on the same subject, with most predictions falling within the 0$\sim$10° range. 

We also conducted localization experiments using the DSN localization model in a cross-subject and inter-subject settings involving three subjects. Fig.~\ref{fig13-2} (a) illustrates the statistical distribution of angular errors, demonstrating that the model's performance exhibits a strong concentration in the 0$\sim$10° range. The results also reveal variations in larger errors, reflecting the challenges of localization when tested on unseen subjects. 
The DSN localization model for inter-subject localization is trained and tested on one volunteer, while the images used for testing are obtained from two other volunteers. The inter-subject localization results are illustrated in Fig.~\ref{fig13-2} (b). The angular errors in Fig.~\ref{fig13-2} (b) are more evenly distributed across higher ranges in inter-subject localization due to anatomical variations. An inverse correlation is observed between SSIM and localization error. Nearly all retrieval results with angular errors below 20° exhibit SSIM values above the median, supporting the reliability of AIA-UltraNeRF's localization, as lower angular discrepancies consistently correspond to higher SSIM between the retrieved and reference images. 

AIA-UltraNeRF provides a reliable initial pose estimate of the standard planes when anatomical features are delineated. To enhance localization robustness, a binarization-based approach that emphasizes alignment with key anatomical regions may reduce the influence of irrelevant background information.

\section{Conclusion}\label{chapter 9}

We designed a RUSS, assisted by the RCM mechanism, utilizing AIA-UltraNeRF to potentially enable the separation of scanning and diagnostic workflows. 
This separation eliminates the need for operator-dependent probe adjustments, allowing sonographers to analyze ultrasound images more efficiently, similar to the workflow in CT imaging.  AIA-UltraNeRF models the 3D ultrasound map as a continuous function to implicitly characterize the color of ultrasound images in terms of acoustic impedance. A dual-supervised network, employing teacher and student models, is presented to hash encode the rendered ultrasound images from the reconstructed map for ultrasound localization. Thus, AIA-UltraNeRF can precisely localize images by leveraging matching hash values, which provide an offline initial image position. Simulations and experiments indicate that AIA-UltraNeRF can effectively characterize the acoustic impedance for ultrasound reconstruction and render new-view ultrasound images for image plane localization. Moreover, AIA-UltraNeRF achieves localization on the phantom and offers a reliable initial pose estimation on human subjects. In the future, we will focus on dynamic scenes, accounting for tissue movement caused by respiratory motion to further enhance its clinical applicability.

%

\bibliographystyle{IEEEtran}
\bibliography{refs}{}

\vfill

\end{document}